\documentclass[10pt,twocolumn,letterpaper]{article}

\usepackage{iccp}
\usepackage{times}
\usepackage{amsmath}
\usepackage{amssymb}
\usepackage{url}
\usepackage{graphicx}
\usepackage{mathtools}
\usepackage{hyperref}
\usepackage{booktabs}
\usepackage{stackengine}
\newcommand\xrowht[2][0]{\addstackgap[.5\dimexpr#2\relax]{\vphantom{#1}}}

\iccpfinalcopy 

\def\iccpPaperID{****} 

\ificcpfinal\fi
\begin{document}

\title{A Comparative Study for Non-rigid Image Registration and Rigid Image Registration\thanks{This is a final project report for EE 239AS Computational Imaging in Fall 2019 at UCLA. The project is finished at a very limited time. 
}}

\author{Xiaoran Zhang\\
UCLA\\
{\tt\small xiaoran108@ucla.edu}
\and
Hexiang Dong\\
UCLA\\
{\tt\small bobdong233@ucla.edu}
\and
Di Gao\\
UCLA\\
{\tt\small di0396@ucla.edu}
\and
Xiao Zhao\\
UCLA\\
{\tt\small zhxiao97@ucla.edu}
}

\maketitle
\thispagestyle{empty}

\begin{abstract}
   Image registration 
   algorithms can be generally categorized into two groups: non-rigid and rigid. Recently, many deep learning-based algorithms employ a neural net to characterize non-rigid image registration function. However, do they always perform better? In this study, we compare the state-of-art deep learning-based non-rigid registration approach with rigid registration approach. The data is generated from Kaggle Dog vs Cat Competition \url{https://www.kaggle.com/c/dogs-vs-cats/} and we test the algorithms' performance on rigid transformation including translation, rotation, scaling, shearing and pixelwise non-rigid transformation. The Voxelmorph is trained on \textit{rigidset} and \textit{nonrigidset} separately for comparison and we also add a gaussian blur layer to its original architecture to improve registration performance. The best quantitative results in both root-mean-square error (RMSE) and mean absolute error (MAE) metrics for rigid registration are produced by SimpleElastix and non-rigid registration by Voxelmorph. We select representative samples for visual assessment.
\end{abstract}

\section{Introduction}

Image registration is a fundamental task in image processing, whose importance soars with growing number of different types of devices and increasing availability. It also serves as a crucial step in a great variety of biomedical imaging applications. Registration techniques can be generally divided into rigid registration and non-rigid registration \cite{oh2017deformable}. Rigid image registration moves or rotates image pixels uniformly so that the pixelwise relations are kept before and after the transformation. Non-rigid image registration method, also known as deformable registration, changes the pixelwise relation and produces a translation map for each pixel. 

The registration task requires user to input a pair of images acquired from different devices, among which one denoted as \textit{moving}, indicating the image that needs to be aligned, and the other denoted as \textit{fixed}, indicating the target coordinate system of the alignment, and outputs the aligned image after transformation, commonly denoted as \textit{moved}. Traditional rigid registration methods often solve an optimization problem for each pair of images. However, solving a pairwise optimization problem can be computationally intensive and slow in practice \cite{balakrishnan2019voxelmorph}. With the advent of deep learning in recent decades, a number of Convolutional Neural Network (CNN) architectures are proposed for registration. The pipelines of deep learning-based techniques consist of two types. One is to use a CNN to directly model the transformation from the input image pair and the final aligned output. The limitation is that it requires ground truth during the training phase, either the pixelwise translation map (registration field) or the aligned image corresponding to the \textit{moving} image. The output \textit{moved} image is not necessarily the same as the \textit{fixed} image in all registration scenarios due to the measurement noise and the artifacts in the generation process of the \textit{fixed} image. Thus, the ground truth is usually hard to acquire. The other type of the pipeline is to use a CNN to model the registration field and utilize a Spatial Transformer Network \cite{jaderberg2015spatial} to perform the registration instead of modeling the transformation directly, making the pipeline unsupervised.

Although there are a number of approaches proposed for image registration, few of them provide a thorough study for the differences between the state-of-art deep learning-based non-rigid registration trained with rigid and non-rigid data and rigid registration approaches on rigid and non-rigid testing data recently. In our work, we compared several state-of-art non-rigid and rigid registration frameworks and our major contributions are: 1) we generated our data set using images from the Kaggle Dog vs Cat Competition; 2) we reproduced the state-of-art 3D unsupervised non-rigid registration approach Voxelmorph \cite{balakrishnan2019voxelmorph} in 2D and improve the registration results by adding a gaussian layer for registration field compared to the original architecture; 3) we reproduced several state-of-art rigid registration methods including SimpleElastix \cite{marstal2016simpleelastix}, Oriented FAST and Rotated BRIEF (ORB) \cite{rublee2011orb} and intensity-based image registration by \href{https://www.mathworks.com/help/images/intensity-based-automatic-image-registration.html#d117e16633}{Matlab}.

\section{Related Work}
\subsection{Rigid image registration}
Rigid image registration generally utilizes a linear transformation, which includes translation, rotation, scaling, shearing and other affine transformations. Extensive studies have been conducted on the topic of rigid registration \cite{eggert1997estimating,letteboer2003rigid,leroy2004rigid,commowick2012block,debayle2016rigid,ourselin2000block,feldmar1996rigid}. Recently, there are three widely used tools for rigid registration. The first is the intensity-based approach \cite{rohde2003adaptive,myronenko2010intensity,klein2009elastix}. Matlab has embedded function called imregister for this method, making it accessible and easy to use. The second is the ORB based approach, which builds on FAST key points detector \cite{rosten2006machine,rosten2008faster} and BRIEF feature descriptor \cite{calonder2010brief}. The third is the famous SimpleElastix model \cite{marstal2016simpleelastix}, which is an extension of Elastix \cite{klein2009elastix}. SimpleElastix also contains spline non-rigid transformation function. These rigid registration methods require user to specify the transformation model before the registration, which limits their generalization ability when dealing when unknown transformation.

\subsection{Non-rigid image registration}
Several studies propose pairwise optimization methods for non-rigid image registration within displacement vector fields, including elastic type models, free-form deformations with b-splines \cite{rueckert1999nonrigid}, discrete methods \cite{dalca2016patch,glocker2008dense} and Demons \cite{pennec1999understanding,thirion1998image}. There are also several methods proposing diffeomorphic transformation-based methods including Large Diffeomorphic Distance Metric Mapping (LDDMM) \cite{beg2005computing,zhang2017frequency,cao2005large,ceritoglu2009multi,hernandez2009registration,joshi2000landmark,oishi2009atlas}, DARTEL \cite{ashburner2007fast} and diffeomorphic demons \cite{vercauteren2009diffeomorphic}. These methods are not learning based and need to be repeated for each pair, which is time-consuming when dealing with large data set.

There are also some recent papers proposing using neural networks to learn the registration function, but all of them rely on ground truth translation map \cite{cao2017deformable,krebs2017robust,rohe2017svf,sokooti2017nonrigid,yang2017quicksilver,liao2017artificial,cao2018deformable}. In common registration applications, it is hard to acquire the translation map from two natural images taken by two different camera systems. Hu \etal \cite{hu2018weakly}  put forward a weakly supervised deep learning-based registration approach, but it still requires a proportion of ground truth. More recently, several unsupervised methods have been proposed \cite{de2017end,li2017non,li2018non}. They utilize neural networks to model the registration field and then apply the spatial transformer network \cite{jaderberg2015spatial} to warp the image. However, the methods are only tested on a small subsets of volumes, such as 3D regions and have not been compared with other popular models like LDDMM or U-net. Balakrishnan \etal \cite{balakrishnan2018unsupervised} then propose an unsupervised learning method for deformable image registration. The group extends the method to Voxelmorph \cite{balakrishnan2019voxelmorph} and demonstrates impressive performance on various data set, which is considered as the state-of-art. In this work, we reproduce this paper and we improve the registration result by adding a gaussian layer after obtaining the registration flow.

\section{Method}
\subsection{2D Voxelmorph}
Let $(I_m,I_f)$ be the input pair for the Voxelmorph over 2D spatial domain $\Omega=\mathbb{R}^2$, where $I_m$ denotes \textit{moving} image and $I_f$ denotes \textit{fixed} image. The Voxelmorph models the registration field function (pixelwise translation map) $g_\theta(I_m,I_f)=\mathbf{\phi}$ by using a neural network. The $\theta$ denotes the network parameter and $\phi$ denotes the estimated registration field. Voxelmorph utilizes a Spatial Transformer Network (STN) \cite{jaderberg2015spatial} to compute the \textit{moved} image $I_m\circ\phi$. Stochastic gradient descent is used to find the optimal $\hat{\theta}$.

\begin{figure}
    \centering
    \includegraphics[scale=0.3]{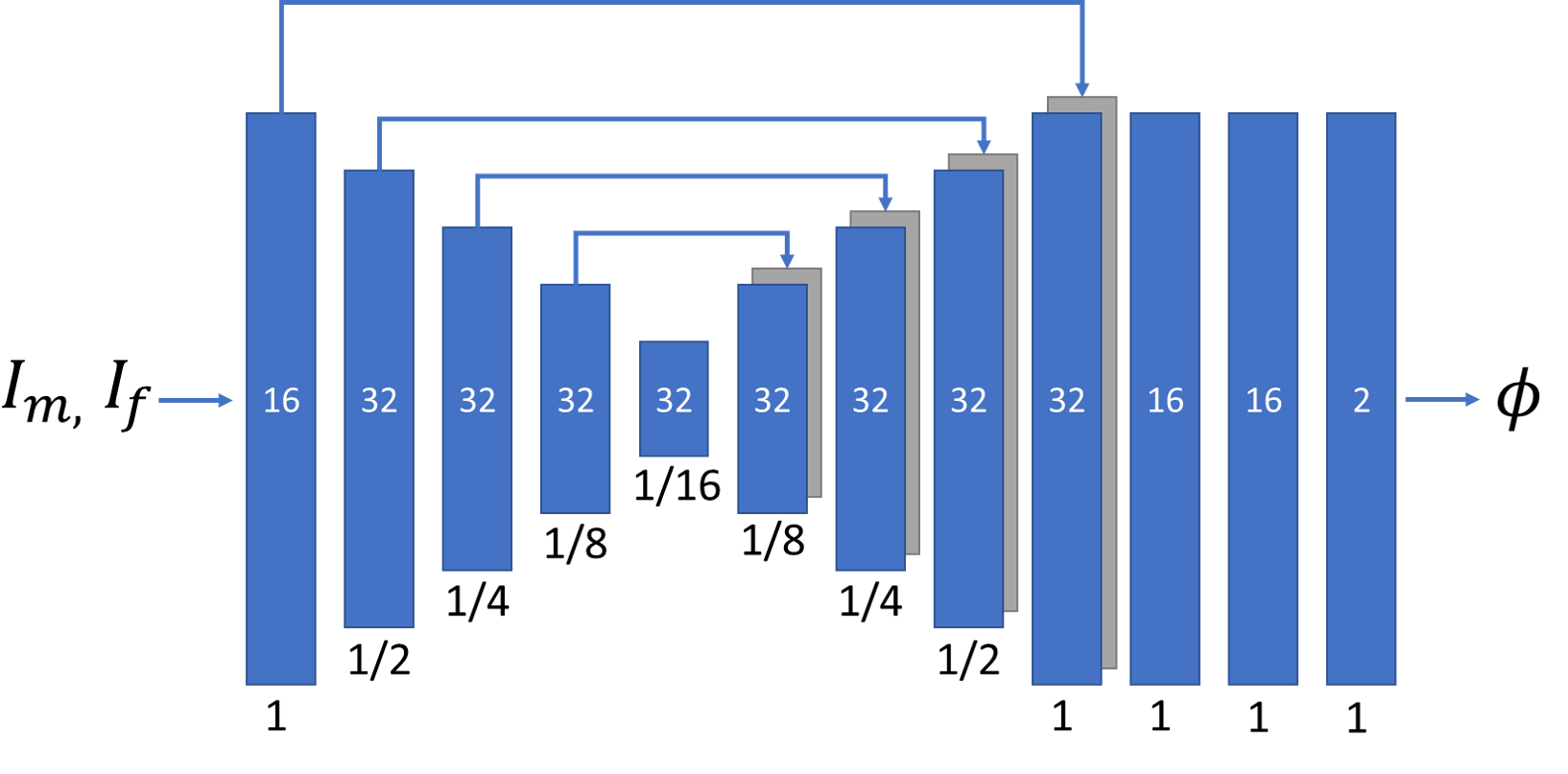}
    \caption{U-net based architecture used to model the registration field.}
    \label{fig:Unet}
\end{figure}

The CNN architecture used in the 2D Voxelmorph is based on U-net. First proposed by Ronneberger \etal \cite{ronneberger2015u} at 2015, U-net architecture has widely been used in registration and segmentation. The architecture implemented in our work is shown in Fig.~\ref{fig:Unet}.
The input size of the U-net architecture is $256\times256\times6$ as we concatenate the RGB channels of the \textit{moving} and \textit{fixed} image. Convolutions in 2D with kernel size 3 and stride 2 are implemented in the encoder and decoder. Each convolution is followed by a LeakyReLU with parameter $0.2$. The original input size is denoted as $1$ for simplification and the successive layer utilize $2\times2$ max pooling operation, shrinking the size by $2$. The size of the output registration field $\phi$ is $256\times256\times2$. Different from the original 3D Voxelmorph architecture, we add a Gaussian blur layer $\phi'=gauss(\phi)$ after registration field to smooth the pixelwise displacement. 

\begin{figure*}
    \centering
    \includegraphics[scale=0.5]{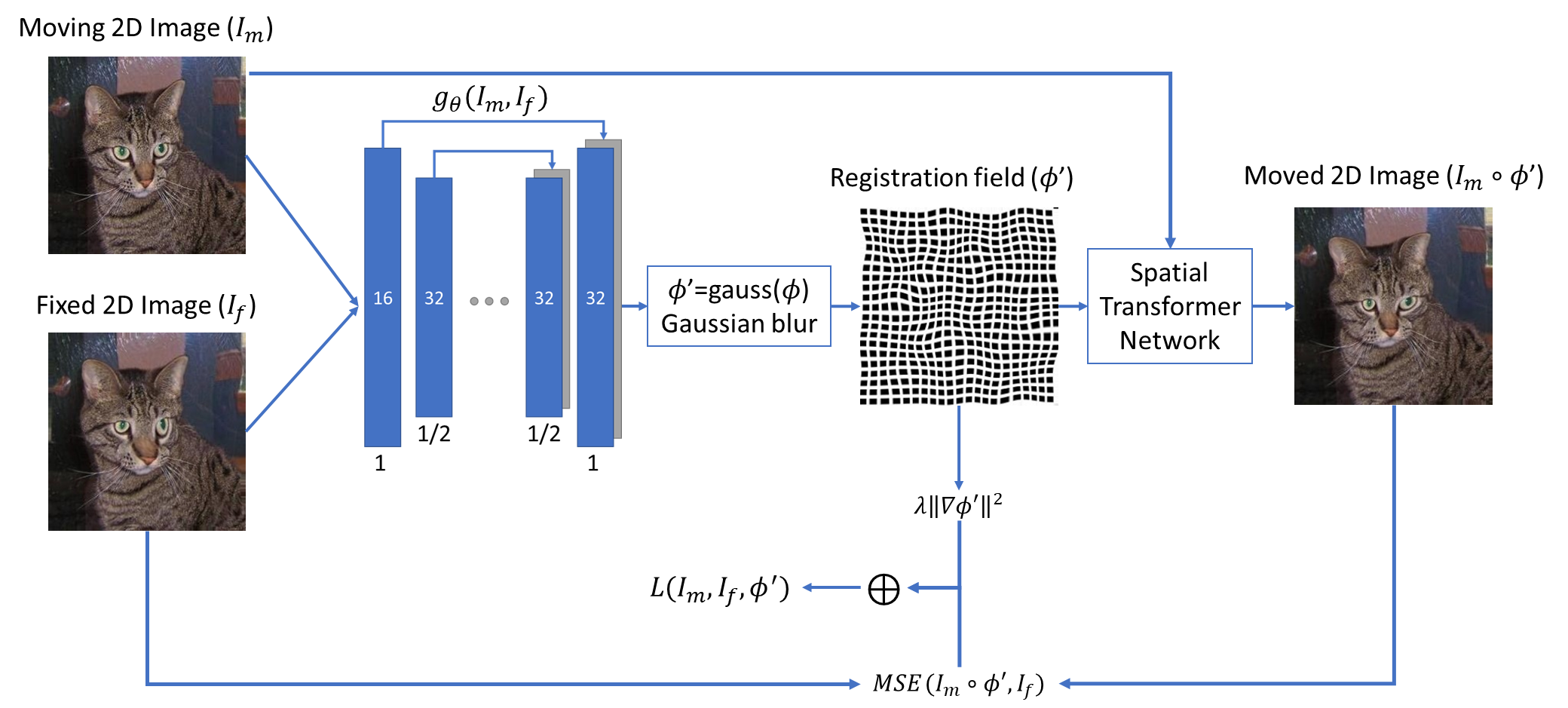}
    \caption{2D Voxelmorph architecture. The input \textit{moving} and \textit{fixed} images are resized to $256\times256\times3$ and concatenated to $256\times256\times6$ for the CNN to estimate registration field. }
    \label{fig:Voxelmorph}
\end{figure*}

After obtaining the registration field, we construct a differentiable operation based on STN \cite{jaderberg2015spatial} to compute the $I_m\circ\phi'$ with bilinear interpolation. We utilize the unsupervised loss function in the original 3D Voxelmorph, which penalizes the appearance difference and local spatial variation in $\phi'$
\begin{equation}
    L(I_m,I_f,\phi')=MSE(I_m,I_f)+\lambda||\Delta\phi'||^2.
\end{equation}
The pipeline of 2D Voxelmorph is depicted in Fig.~\ref{fig:Voxelmorph}.

\subsection{SimpleElastix}
Developed based on Elastix \cite{klein2009elastix}, SimpleElastix is one of the favorite tools for rigid image registration. It also contains non-rigid image library using B-spline polynomial. The main idea of SimpleElastix is to solve a pairwise optimization problem by minimizing the cost function $C$. The optimization can be formulated as
\begin{equation}
    \hat{T} = \underset{T}{\text{argmin}}\; C(T,I_f,I_m),
\end{equation}
with cost function defined as
\begin{equation}
    C(T,I_f,I_m) = -S(T,I_f,I_m)+\gamma P(T),
\end{equation}
where $T$ is the transformation matrix, $S$ is the similarity measurement and $P$ is the penalty term with regularizer parameter $\gamma$. SimpleElastix is based on the parametric approach to solve the optimization problem, where the number of possible transformation is limited by introducing a parametrization (model) of the transform. The optimization becomes
\begin{equation}
    \hat{T}_\mu = \underset{T_\mu}{\text{argmin}}\; C(T_\mu,I_f,I_m),
\end{equation}
$T_\mu$ denotes the parametrization model and vector $\mu$ contains the values of the transformation parameters. In our case for 2D rigid transformation, the parameter vector $\mu$ contains one rotation angle and the translation in $x$ and $y$ direction
\begin{equation}
    \hat{\mu} = \text{argmin}\;C(\mu,I_f,I_m).
\end{equation}

\subsection{ORB}
\begin{figure}
    \centering
    \includegraphics[scale=0.35]{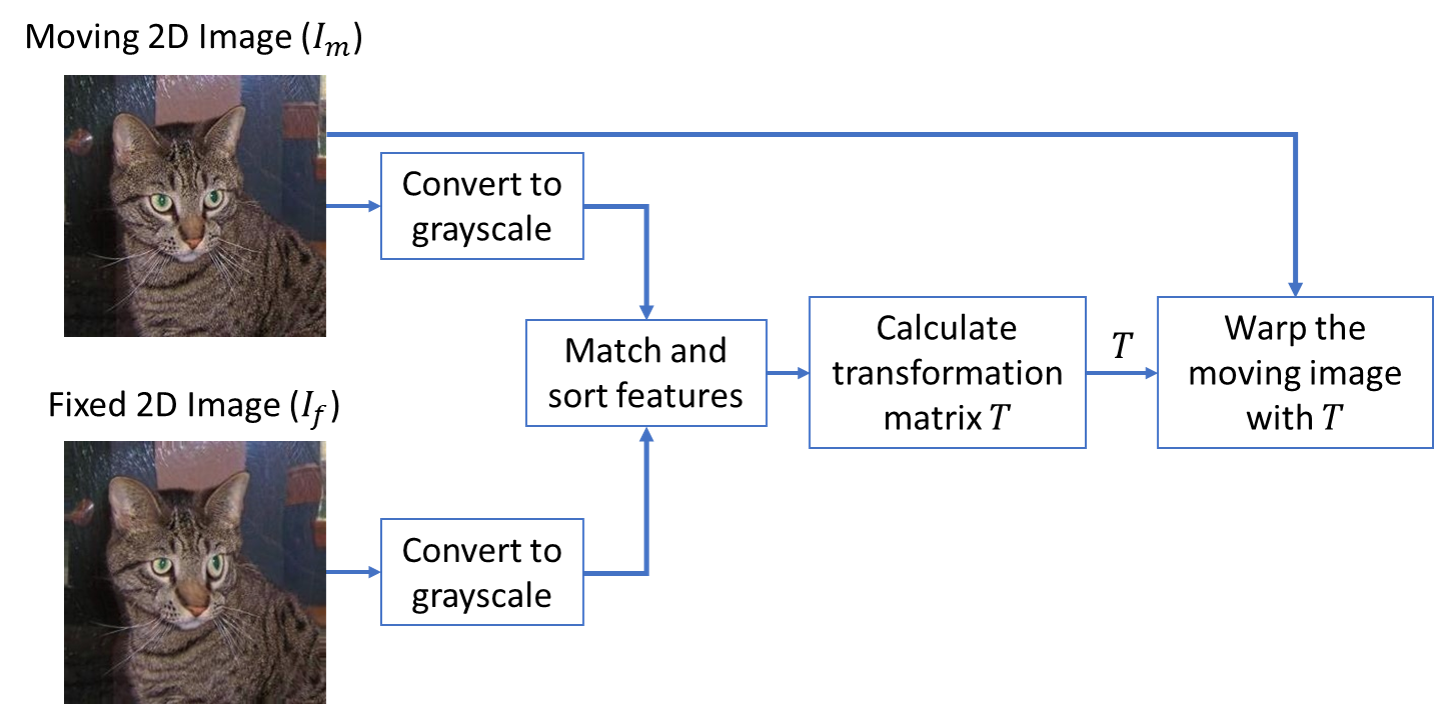}
    \caption{Implementation pipeline for ORB approach.}
    \label{fig:ORB}
\end{figure}
ORB-based registration is often called feature-based, since sparse sets of features are detected and matched in two images \cite{rublee2011orb}. The final output of the method is a \textit{moved} image after calculating the $3\times3$ transformation matrix $T$. The ORB registration approach could be divided into 4 stages including image preprocessing, feature detection, feature matching and image warping. The pipeline is shown in Fig.~\ref{fig:ORB}. We first read both \textit{moving} and \textit{fixed} image and convert them into grayscale using the following empirical function
\begin{equation}
    I_g = 0.299R+0.587G+0.114B
\end{equation}
where $(R,G,B)$ denotes the original pixel value for three channels in $I_m$,$I_f$ and $I_g$ denotes the output grayscale image. We then use feature detector, which consists of a locator and descriptor, to extract features from the input image. A locator identifies points on the image that are consistent under image transformations and a detector tells the appearance of the identified points by encoding them into arrays of numbers. In the implementation, we adopt FAST locator and BRIEF descriptor. In the next stage, we match the generated features using hamming distance and sort out the top corresponding points in the two images for the transformation matrix calculation. RANSAC is further utilized for improving the robustness. In the last stage, we warp the image with the $T$ to calculate the final output.

\subsection{Intensity-based Registration}
\begin{figure}
    \centering
    \includegraphics[scale=0.32]{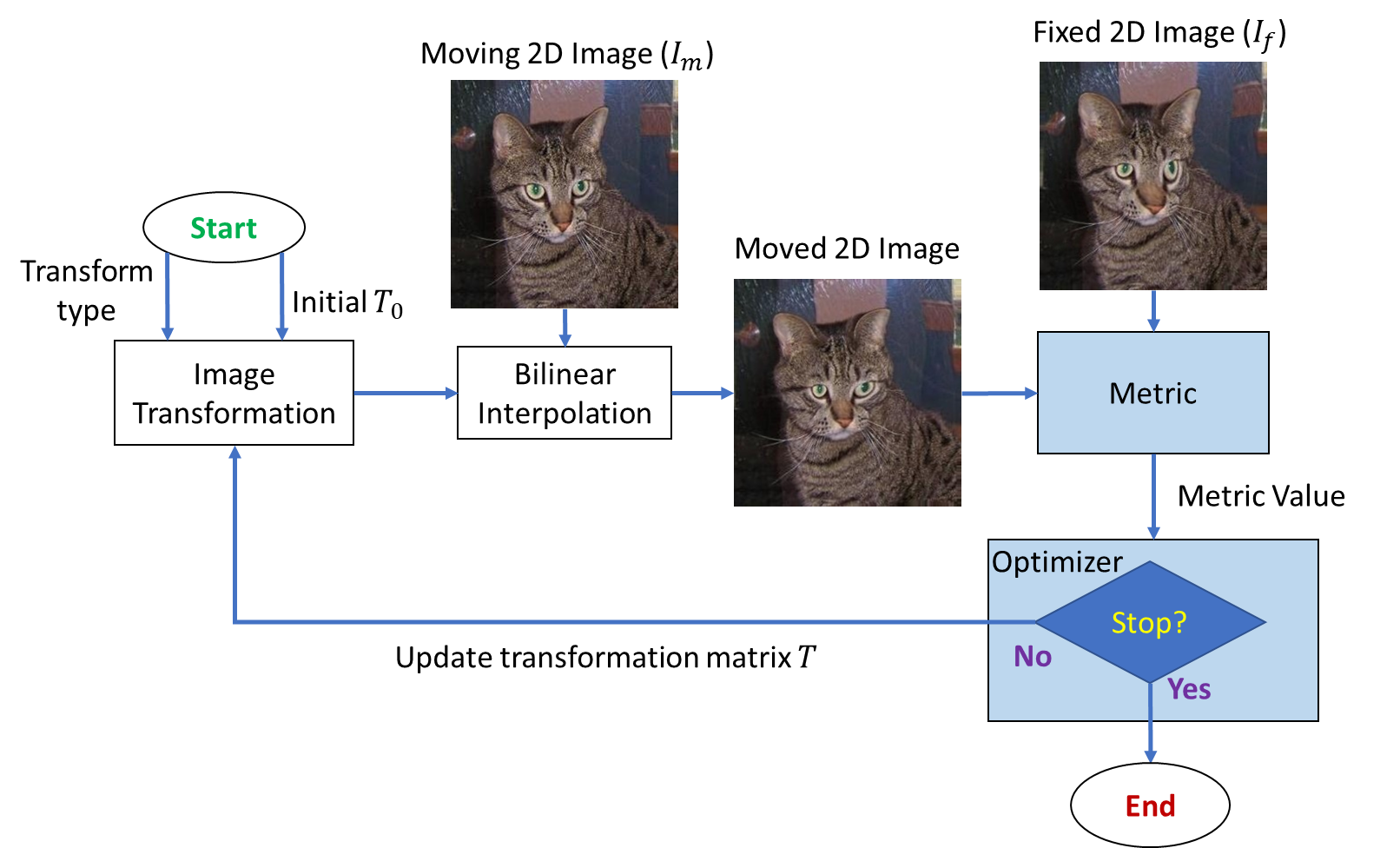}
    \caption{The flow chart for intensity-based image registration using Matlab.}
    \label{fig:Intensity}
\end{figure}

Intensity-based image registration is an iterative optimization process and is widely used in Matlab. The method requires prior information of initial transformation matrix $T_0$, the metric and an optimizer. In our work, we choose mean square similariy metric and regular step gradient descent optimizer. The initial transformation matrix defines the type of 2-D transformation that aligns $I_m$ with $I_f$. The metric is used to describe the similarity for evaluating the accuracy of our registration. We need two images as the input and get a scalar result to show how similar these two images are. And in order to reshape the metric, the optimizer is used to define the method for minimizing or maximizing the similarity metric. The pipeline is shown in Fig.~\ref{fig:Intensity}. The key pairwise optimization objective is to provide accurate estimation on transformation matrix $T$, which is a rigid registration approach.

\section{Experiment}
\subsection{Data generator}
\begin{figure}
    \centering
    \includegraphics[scale=0.4]{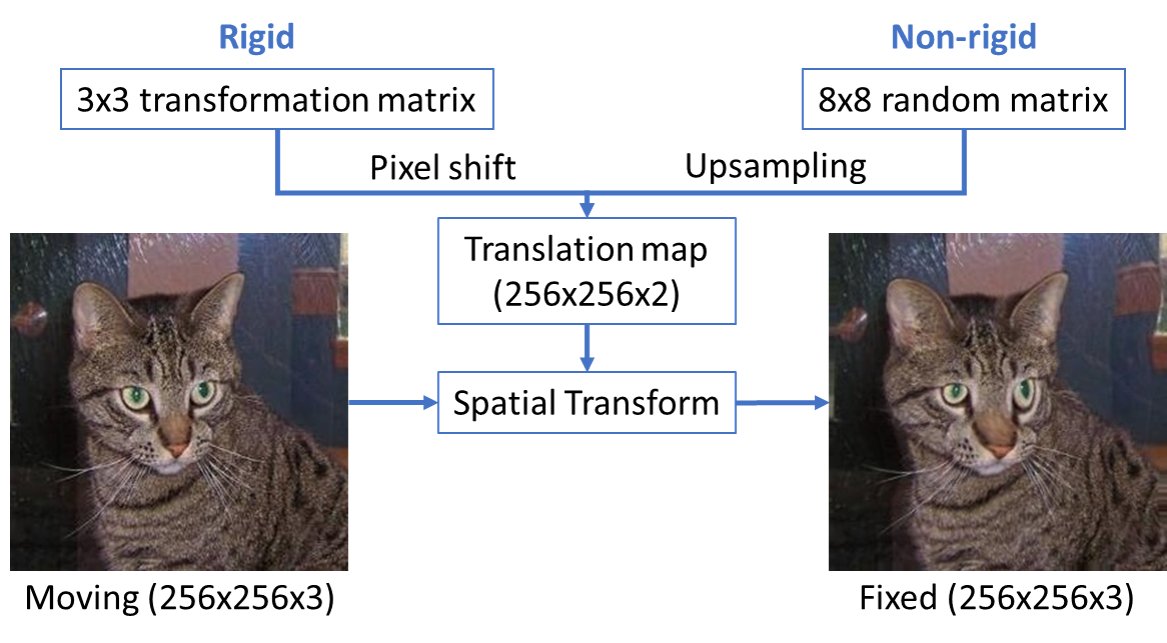}
    \caption{Data generator for \textit{fixed} image from \textit{moving} image.}
    \label{fig:Datagen}
\end{figure}

The data set used in this work is generated from the Kaggle Dogs vs Cats competition. We downloaded 1200 images and separate them into two groups: 1000 images for training and 200 images for testing. These downloaded images are considered as \textit{moving} images. The \textit{fixed} images in the training and testing set are generated using Spatial Transformer Network \cite{jaderberg2015spatial} with ground truth translation map. The pipeline is shown in Fig.~\ref{fig:Datagen}.
The transformation matrices and its random entries for each pair used in the generator are listed in Table~\ref{tab:transmax}.

\begin{table}
\caption{Transformation and random matrix used in the data generator.}
\resizebox{0.5\textwidth}{!}{
\begin{tabular}{|c|c|c|}
\hline
Type & Matrix & Random Seed \\ \hline
 \xrowht{40pt}
Translation  & $\begin{bmatrix}
    1 & 0 & 0 \\
    0 & 1 & 0 \\
    t_x & t_y & 1
\end{bmatrix}$ & $t_x,t_y \in[-5,5]$ \\ \hline \xrowht{40pt}
 Shearing & $\begin{bmatrix}
    1 & sh_x & 0 \\
    sh_y & 1 & 0 \\
    0 & 0 & 1
\end{bmatrix}$ & $sh_x,sh_y \in[0,0.15]$ \\ \hline \xrowht{40pt}
 Scaling & $\begin{bmatrix}
    s_x & 0 & 0 \\
    0 & s_y & 0 \\
    0 & 0 & 1
\end{bmatrix}$ & $s_x,s_y \in[0.9,1]$ \\ \hline \xrowht{40pt}
 Rotation & $\begin{bmatrix}
    \cos(q) & \sin(q) & 0 \\
    -\sin(q) & \cos(q) & 0 \\
    0 & 0 & 1
\end{bmatrix}$ & $q\in[-5,5]$ \\\hline \xrowht{40pt}
 Pixelwise& $\begin{bmatrix}
    p_{11} & ... & p_{18} \\
     \vdotswithin{p_{81}} & \vdotswithin{...} & \vdotswithin{p_{88}} \\
    p_{81} & ... & p_{88}
\end{bmatrix}$ & $p_{ij} \in[-5,5]$ \\ \hline 
\end{tabular}} \label{tab:transmax}
\end{table}
The rigid transformation matrix $T$ is a $3\times3$ matrix. We take the pixel shift in Cartesian coordinate system to calculate the translation map from $T$. Let $[x,y,1]^T$ denote the homogeneous coordinate in \textit{moving} image and $[x',y',1]^T$ denote the coordinate in the \textit{fixed} image, we have
\begin{equation}
    \begin{bmatrix}
        x' \\ y' \\1
\end{bmatrix}=\begin{bmatrix}
       t_{11} & t_{12} & t_{13}\\
       t_{21} & t_{22} & t_{23}\\
       t_{31} & t_{32} & t_{33}
\end{bmatrix}
\begin{bmatrix}
        x \\ y \\1
\end{bmatrix},
\end{equation}
and pixel shift can be calculated as
\begin{equation}
    \begin{bmatrix}
            \Delta x\\
            \Delta y
    \end{bmatrix}
    =    \begin{bmatrix}
            x'-x\\
            y'-y
    \end{bmatrix}.
\end{equation}
Thus we could have the ground truth translation map with shape $256\times256\times2$, where the first channel represents the pixel shift in $x$ and second represents $y$ for each pixel. For non-rigid transformation, we produce a $8\times8$ random matrix and upsample it to $256\times256$ for $x$ and $y$ and concatenate the two channels to generate the ground truth translation map. Random seed in Table~\ref{tab:transmax} denotes the random entries generated in the transformation matrices for each image pair. For instance, in translation transformation, the random entries in $T$ are $t_x$ and $t_y$ in range $[-5,5]$. 

Two different types of training data are produced using the technique described above. The first type, \textit{rigidset}, is generated by separating our 1000 downloaded images into 5 categories, each containing 200 images, and conducting 
spatial transformations mentioned in Table~\ref{tab:transmax} on each category separately. The second type,  \textit{nonrigidset}, is generated with the entire 1000 images using translation map upsampled from the pixelwise random matrix. The testing set is separated into 5 types, which are translation, rotation, scaling, shearing, pixelwise nonrigid, and each  contains 40 images. We test the performance of Voxelmorph using \textit{rigidset} and \textit{nonrigidset} separately and compare the result. For notation simplicity, we denote Voxelmorph(NN) as trained with \textit{nonrigidset} without gaussian layer, Voxelmorph(RN) as trained with \textit{rigidset} without gaussian layer, Voxelmorph(NG) as trained with \textit{nonrigidset} with gaussian layer and Voxelmorph(RG) as trained with \textit{rigidset} with gaussian layer in the following article.

\subsection{Experiment setup}
The Voxelmorph is implemented in Python with Keras in Tensorflow backend and CUDA Deep Neural Network (cuDNN) library. The model is trained and tested on NVIDIA GPU GTX 2080 Ti with 11GB memory. The total number of epoch is 1500 and each image is resized to $256\times256\times 3$. The SimpleElastix, ORB are implemented in Python and Intensity-based registration is implemented in Matlab.

\subsection{Evaluation metrics}
The quantitative evaluation is conducted by calculating root-mean-square error (RMSE) and mean absolute error (MAE) between the estimated translation map and ground truth translation map, each with a size of $256\times256\times2$. Two channels represent pixel shift in $x$ and $y$ separately.

\subsubsection{Root mean square error}
Let $\hat{t}_{ij}$ denotes the element in estimated translation map and $t_{ij}$ denotes the element in ground truth translation map. The RMSE is calculated as
\begin{equation}
    RMSE = \sqrt{\frac{1}{N}\sum_{j=1}^{N_{col}}\sum_{i=1}^{N_{row}}(\hat{t}_{ij}-t_{ij})^2},
\end{equation}
where $N$ denotes the total number of points, $N_{col}$ denotes number of column pixels, $N_{row}$ denotes number of row pixels. In our case, $N_{col}=N_{row}=256$.

\subsubsection{Mean absolute error}
The MAE is calculated as
\begin{equation}
    MAE = \frac{1}{N}\sum_{j=1}^{N_{col}}\sum_{i=1}^{N_{row}}|\hat{t}_{ij}-t_{ij}|.
\end{equation}


\section{Results}
\subsection{Quantitative assessment}
The quantitative assessment using RMSE metric in Cartesian $x$ and $y$ are reported in Table~\ref{tab:RMSEx} and Table~\ref{tab:RMSEy} respectively. The MAE in $x$ and $y$ are reported in Table~\ref{tab:MAEx} and Table~\ref{tab:MAEy}. The best performance observed in rigid transformation testing is implemented by SimpleElastix. It achieves a high score in both RMSE and MAE with an average of $0.11$ in $x$, $0.11$ in $y$ reported in RMSE and $0.09$ in $x$, $0.09$ in y reported in MAE. In non-rigid transformation, Voxelmorph(RN) achieves the best score with $2.63$ in $x$ and $2.63$ in $y$ using RMSE metric, $2.16$ in $x$ and $2.14$ in $y$ using MAE metric.

From Table~\ref{tab:RMSEx}, we notice that by introducing a gaussian blur layer, Voxelmorph(RG) improves the RMSE score significantly compared with Voxelmorph(RN) in scaling and shearing, which is the original architecture trained with \textit{rigidset}. Similar results are also observed in Table~\ref{tab:RMSEy}, Table~\ref{tab:MAEx} and Table~\ref{tab:MAEy}. This demonstrates the effectiveness of the gaussian blur, which smooths the translation map. In our tasks such as translation, rotation and non-rigid pixelwise transformation, Voxelmorph with gaussian blur layer shows relative the same result after training using both \textit{nonrigidset} and \textit{rigidset}.

\begin{table*}[htbp]
\caption{RMSE error for x coordinate in pixel(px).}
\resizebox{1\textwidth}{!}{
\begin{tabular}{cccccccc}
\toprule
RMSE(px)        & SimpleElastix & ORB       & Intensity-based & Voxelmorph(NN) & Voxelmorph(RN) & Voxelmorph(NG) & Voxelmorph(RG) \\\midrule

Translation & $\mathbf{0.11\pm 0.08}$     & $0.26\pm 0.21$ & $0.28\pm 0.18$       & $3.25\pm 1.26$      & $3.03\pm 1.29$      & $3.58\pm 1.12$      & $3.9\pm 2.00$       \\

Rotation    & $\mathbf{0.13\pm 0.09}$     & $0.31\pm 0.23$ & $0.28\pm 0.13$       & $6.88\pm 3.24$      & $7.06\pm 3.07$      & $7.06\pm 3.14$      & $8.33\pm 3.73$      \\

Scaling     & $\mathbf{0.09\pm 0.08}$     & $0.48\pm 0.66$ & $1.11\pm 1.21$       & $6.44\pm 4.15$      & $7.00\pm 4.16$      & $6.83\pm 3.98$      & $\mathit{6.26\pm 3.52}$      \\

Shearing    & $\mathbf{0.11\pm 0.10}$     & $0.45\pm 0.31$ & $4.80\pm 3.20$       & $11.26\pm 6.4$      & $10.18\pm 5.95$     & $11.75\pm 6.57$     & $\mathit{6.63\pm 3.64}$      \\

Pixelwise   & $3.91\pm 3.60$     & $4.60\pm 3.40$ & $2.83\pm 1.98$       & $3.00\pm 0.09$      & $\mathbf{2.63\pm 0.07}$      & $3.23\pm 0.14$      & $2.87\pm 0.18$\\   
\bottomrule    

\end{tabular}}\label{tab:RMSEx}
\end{table*}

\begin{table*}[htbp]
\caption{RMSE error for y coordinate in pixel(px).}
\resizebox{1\textwidth}{!}{
\begin{tabular}{cccccccc}
\toprule
RMSE(px)        & SimpleElastix & ORB       & Intensity-based & Voxelmorph(NN) & Voxelmorph(RN) & Voxelmorph(NG) & Voxelmorph(RG) \\\midrule

Translation & $\mathbf{0.10\pm 0.10}$     & $0.26\pm 0.21$ & $0.30\pm 0.20$       & $2.94\pm 1.47$      & $2.54\pm 1.55$      & $3.24\pm 1.36$      & $2.76\pm 1.81$      \\

Rotation    & $\mathbf{0.11\pm 0.10}$     & $0.29\pm 0.26$ & $0.26\pm 0.13$       & $6.93\pm 3.25$      & $7.14\pm 3.18$      & $7.20\pm 2.94$      & $8.23\pm 3.93$      \\

Scaling     & $\mathbf{0.11\pm 0.08}$     & $0.47\pm 0.56$ & $1.78\pm 1.58$       & $6.91\pm 3.87$      & $6.27\pm 3.56$      & $\mathit{6.22\pm 3.38}$      & $\mathit{5.60\pm 3.18}$      \\

Shearing    & $\mathbf{0.13\pm 0.13}$     & $0.43\pm 0.34$ & $7.80\pm 4.43$       & $9.46\pm 5.85$      & $8.95\pm 5.36$      & $10.21\pm 5.93$     & $\mathit{6.31\pm 3.87}$      \\

Pixelwise   & $3.87\pm 2.27$     & $4.24\pm 2.64$ & $3.23\pm 2.31$       & $2.89\pm 0.16$      & $\mathbf{2.63\pm 0.08}$      & $3.19\pm 0.19$      & $2.90\pm 0.17$\\   
\bottomrule
\end{tabular}}\label{tab:RMSEy}
\end{table*}

\begin{table*}[htbp]
\caption{MAE error for x coordinate in pixel(px).}
\resizebox{1\textwidth}{!}{
\begin{tabular}{cccccccc}
\toprule
MAE(px)          & SimpleElastix & ORB       & Intensity-based & Voxelmorph(NN) & Voxelmorph(RN) & Voxelmorph(NG) & Voxelmorph(RG) \\ \midrule

Translation & $\mathbf{0.09\pm 0.07}$     & $0.21\pm 0.18$ & $0.23\pm 0.15$       & $2.98\pm 1.31$      & $2.90\pm 1.32$      & $3.15\pm 1.14$      & $3.75\pm 2.02$      \\

Rotation    & $\mathbf{0.11\pm 0.07}$     & $0.26\pm 0.19$ & $0.23\pm 0.10$       & $5.95\pm 2.91$      & $5.94\pm 2.68$      & $6.05\pm 2.85$      & $6.94\pm 3.23$      \\

Scaling     & $\mathbf{0.08\pm 0.07}$     & $0.41\pm 0.55$ & $0.93\pm 1.02$       & $5.35\pm 3.52$      & $5.90\pm 3.57$      & $5.62\pm 3.43$      & $\mathit{5.09\pm 3.06}$      \\

Shearing    & $\mathbf{0.09\pm 0.09}$     & $0.37\pm 0.26$ & $3.99\pm 2.65$       & $9.92\pm 5.63$      & $8.53\pm 5.03$      & $10.28\pm 5.92$     & $\mathit{4.57\pm 2.50}$      \\

Pixelwise   & $3.23\pm 2.95$     & $3.84\pm 2.83$ & $2.36\pm 1.62$       & $2.41\pm 0.07$      & $\mathbf{2.16\pm 0.05}$      & $2.56\pm 0.12$      & $2.29\pm 0.14$\\\bottomrule    
\end{tabular}}\label{tab:MAEx}
\end{table*}

\begin{table*}[htbp]
\caption{MAE error for y coordinate in pixel(px).}
\resizebox{1\textwidth}{!}{
\begin{tabular}{cccccccc}
\toprule
MAE(px)          & SimpleElastix & ORB       & Intensity-based & Voxelmorph(NN) & Voxelmorph(RN) & Voxelmorph(NG) & Voxelmorph(RG) \\\midrule

Translation & $\mathbf{0.09\pm 0.08}$     & $0.22\pm 0.17$ & $0.25\pm 0.17$       & $2.64\pm 1.49$      & $2.39\pm 1.58$      & $2.83\pm 1.37$      & $2.58\pm 1.86$      \\

Rotation    & $\mathbf{0.09\pm 0.08}$     & $0.24\pm 0.21$ & $0.23\pm 0.11$       & $5.98\pm 2.90$      & $6.02\pm 2.86$      & $6.22\pm 2.65$      & $6.89\pm 3.46$      \\

Scaling     & $\mathbf{0.10\pm 0.07}$     & $0.40\pm 0.47$ & $1.51\pm 1.34$       & $5.75\pm 3.32$      & $5.20\pm 2.98$      & $\mathit{5.17\pm 2.96}$      & $\mathit{4.54\pm 2.73}$      \\

Shearing    & $\mathbf{0.11\pm 0.11}$     & $0.36\pm 0.28$ & $6.58\pm 3.78$       & $7.96\pm 5.02$      & $7.40\pm 4.50$      & $8.65\pm 5.20$      & $\mathit{4.53\pm 2.73}$      \\

Pixelwise   & $3.22\pm 1.91$     & $3.53\pm 2.17$ & $2.70\pm 1.90$       & $2.29\pm 0.12$      & $\mathbf{2.14\pm 0.06}$      & $2.58\pm 0.16$      & $2.34\pm 0.13$    \\\bottomrule

\end{tabular}}\label{tab:MAEy}
\end{table*}

\subsection{Visual assessment}
\begin{figure*}
    \centering
    \includegraphics[scale=0.4]{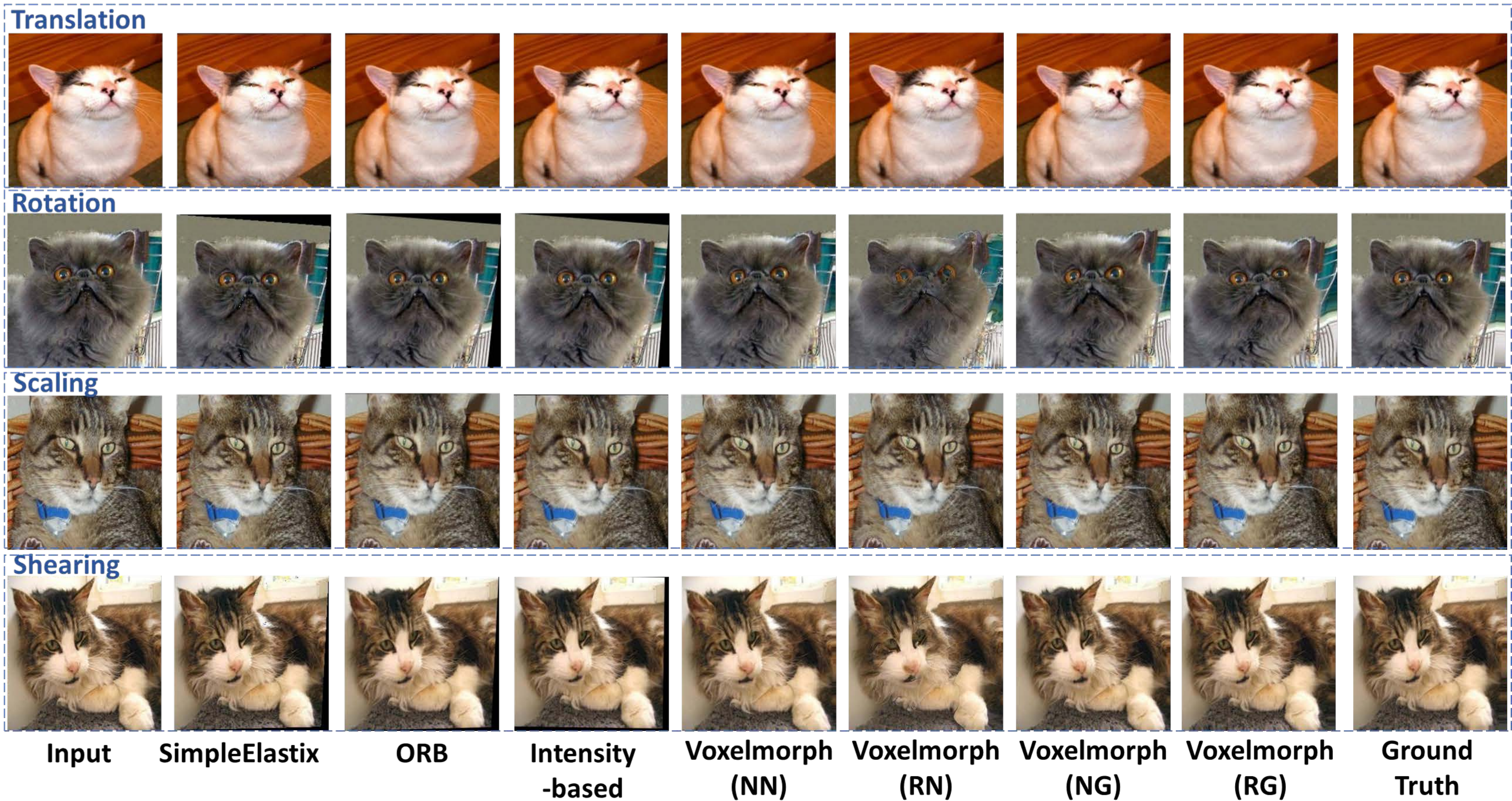}
    \caption{Visual assessment for testing on \textit{rigidset}.}
    \label{fig:rigidtest}
\end{figure*}

\begin{figure*}
    \centering
    \includegraphics[scale=0.4]{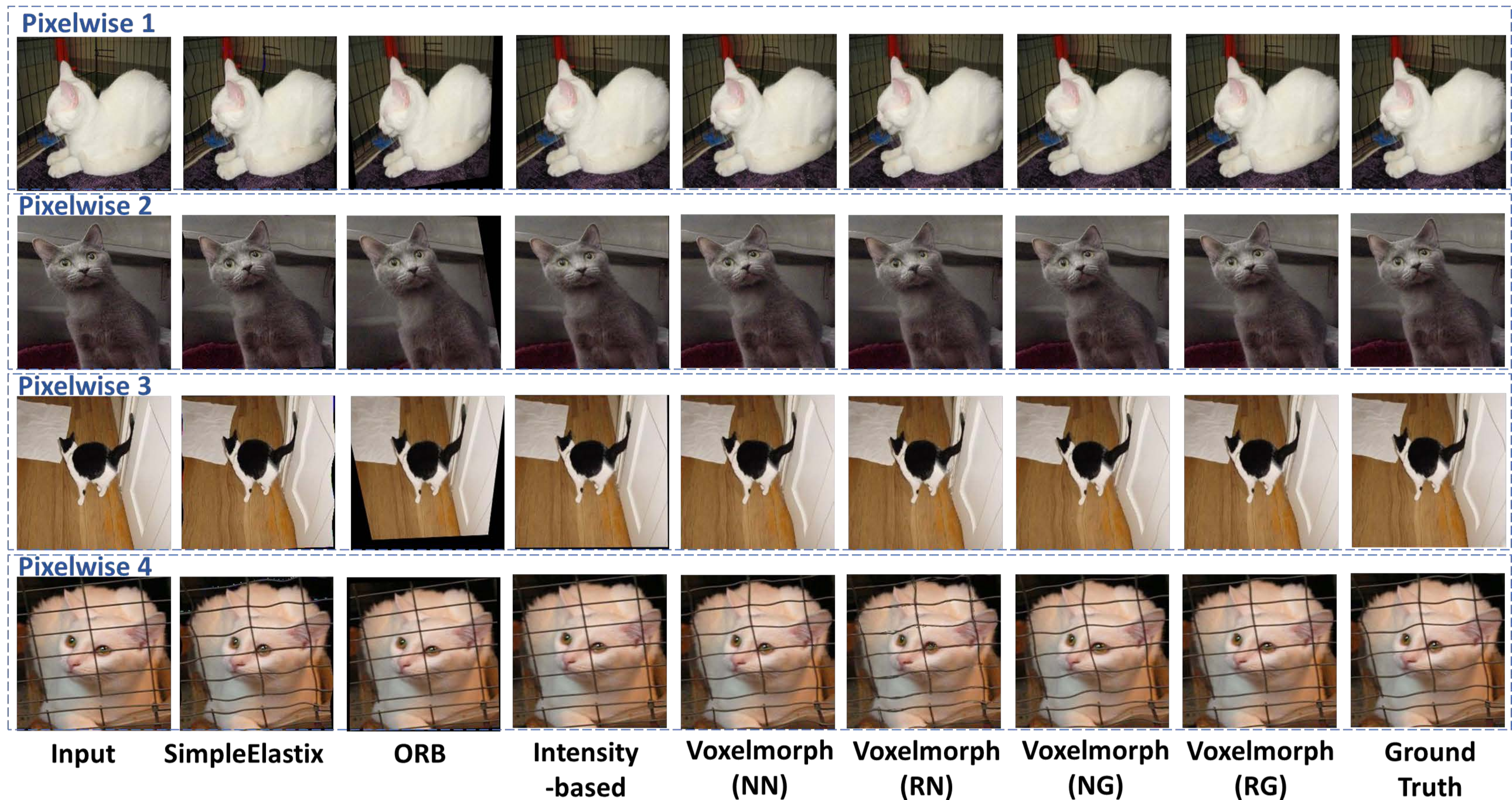}
    \caption{Visual assessment for testing on \textit{nonrigidset}.}
    \label{fig:nonrigidtest}
\end{figure*}
The visual assessment is demonstrated in Fig.~\ref{fig:rigidtest} and Fig.~\ref{fig:nonrigidtest}. We compare the results generated by algorithms mentioned in this paper with the ground truth. In our case, the input is the \textit{moving} image and the ground truth image is the image warped by the ground truth translation map using spatial transformer network. Pixelwise 1-4 in Fig.~\ref{fig:nonrigidtest} denote different types of translation map generated by different random matrices.

From Fig.~\ref{fig:rigidtest}, we can see that rigid transformation methods based on SimpleElastix, ORB and Intensity-based produce a black boundary on \textit{moved} images and lose some information. The reason is that the warping is performed for the entire image instead of each pixel. In column 5-8, Voxelmorph produces a more consistent result compared with column 1-4. We also notice that the training data demonstrates a difference in Voxelmorph. When trained with \textit{rigidset}, Voxelmorph preserves the relative pixel relation better compared to the model trained with \textit{nonrigidset}. For instance, in the shearing row, Voxelmorph(RN) and Voxelmorph(RG) preserve a straight line in a cat body while the Voxelmorph(NN) and Voxelmorph(NG) warp the line into a curve.

From Fig.~\ref{fig:nonrigidtest}, we can see that ORB and Intensity-based approach fail to produce pixelwise \textit{moved} image. For instance in Pixelwise 4, the box lines are still straight in these two methods as the rigid transformation considers a linear transformation instead of a pixelwise warping. SimpleElastix demonstrates a impressive result in non-rigid transformation. Voxelmorph trained in \textit{nonrigidset} and \textit{rigidset} show comparable performance. 

\section{Conclusion}
In this paper, we provide a comparative study for the state-of-art non-rigid image registration method and rigid image registration methods and show that the deep learning-based method doesn's always have a better performance. We reproduce Voxelmorph and its variations from \textit{rigidset} and \textit{nonrigidset}. We also reproduce several rigid transformation approaches including SimpleElastix, ORB and Intensity-based registration. We add a gaussian blur layer and improve the Voxelmorph performance in rigid transformation. Our result is evaluated in terms of RMSE and MAE and it is observed that SimpleElastix demonstrates the best performance in rigid transformation while Voxelmorph(RN) achieves best score in pixelwise transformation. In the future, we intend to evaluate our idea on natural images and combine the advantages of SimpleElastix and Voxelmorph. 





\section{Acknowledgement}
The authors would like to thank Prof. Kadambi and TA Guangyuan Zhao for the excellent teaching and service in the entire quarter.

{\small
\bibliographystyle{ieee}
\bibliography{iccppaper_final}
}

\end{document}